\documentclass[twoside,11pt]{article}
\usepackage[pdftex]{graphicx}
\usepackage{jmlr2e}
\usepackage{url}
\usepackage{algorithm,algorithmic}
\usepackage{amsmath}
\usepackage[small]{caption}
\newcommand{\order}{\ensuremath{\mathcal{O}}}
\newcommand{\tell}{\ensuremath{\tilde{\ell}}}

\newcommand{\w}{\mathbf{w}}
\newcommand{\G}{\mathbf{G}}
\newcommand{\x}{\mathbf{x}}
\newcommand{\g}{\mathbf{g}}
\newcommand{\R}{\mathbb{R}}
\newcommand{\tr}{{\!\top}}
\newcommand{\E}{\ensuremath{\mathbb{E}}}
\newcommand{\Exy}{\ensuremath{\mathbb{E}_{\x,y}}}
\newcommand{\BigBracks}[1]{\Bigl[#1\Bigr]}

\usepackage{xcolor}
\newcommand{\revised}[1]{#1}%{\color{red}#1}}

\begin{document}

\jmlrheading{?}{2012}{?}{?}{?}{A. Agarwal, O. Chapelle. M. Dud\'ik, J. Langford}
\ShortHeadings{A Reliable Effective Terascale Linear Learning System}{A. Agarwal, O. Chapelle. M. Dud\'ik, J. Langford}
\editor{}

\title{A Reliable Effective Terascale Linear Learning System}
\author{%
Alekh Agarwal\thanks{Work was done while all authors were part of Yahoo!\ Research.}
\email{alekha@microsoft.com} \\
\addr{Microsoft Research} \\
\addr{New York, NY}
\AND
Olivier Chapelle \email{olivier@chapelle.cc} \\
\addr{Criteo} \\
\addr{Palo Alto, CA}
\AND
 Miroslav Dud\'ik \email{mdudik@microsoft.com} \\
\addr{Microsoft Research} \\
\addr{New York, NY}
\AND
John Langford \email{jcl@microsoft.com} \\
\addr{Microsoft Research} \\
\addr{New York, NY}
}

\maketitle

\begin{abstract}
We present a system and a set of techniques for learning linear
predictors with convex losses on terascale datasets, with trillions of
features,\footnote{The number of features here refers to the number of
  non-zero entries in the data matrix.} billions of training examples
and millions of parameters in an hour using a cluster of 1000
machines.  Individually none of the component techniques are new, but
the careful synthesis required to obtain an efficient implementation
is.  The result is, up to our knowledge, the most
scalable and efficient linear learning system reported in the
literature (as of 2011 when our experiments were conducted).  We
describe and thoroughly evaluate the components of the system, showing
the importance of the various design choices.
\end{abstract}

\section{Introduction}

Distributed machine learning is a research area that has seen a
growing body of literature in recent years.  Much work focuses on
problems of the form
\begin{equation}
  \min_{\w \in \R^d}~~\sum_{i=1}^n\ell(\w^\tr\x_i;\,y_i) + \lambda
  R(\w),
  \label{eqn:objective}
\end{equation}
where $\x_i$ is the feature vector of the $i$-th example, $y_i$ is the
label, $\w$ is the linear predictor, $\ell$ is a loss function and $R$
is a regularizer. Most distributed methods for optimizing the
objective~\eqref{eqn:objective} exploit its natural decomposability
over examples, partitioning the examples over different nodes in a
distributed environment such as a cluster.

Perhaps the simplest strategy when the number of examples $n$
is too large for a given learning algorithm is to reduce the dataset
size by subsampling. However, this strategy only works if the
problem is simple enough or the number of parameters is very
small. The setting of interest here is when a large number of
examples is really needed to learn a good model.
Distributed algorithms are a natural choice for such scenarios.

It might be argued that even for these large problems, it is
more desirable to explore multicore solutions developed for single
machines with large amounts of fast storage and memory, rather than a
fully distributed algorithm which brings additional complexities due
to the need for communication over a network. Yet, we claim that
there are natural reasons for studying distributed machine learning on
a cluster. In many industry-scale applications, the datasets
themselves are collected and stored in a decentralized fashion over a
cluster, typical examples being logs of user clicks or search
queries. When the data storage is distributed, it is
much more desirable to also process it in a distributed fashion to
avoid the bottleneck of data transfer to a single powerful
server. Second, it is often relatively easy to get access to a
distributed computing platform such as Amazon EC2, as opposed to
procuring a sufficiently powerful server. Finally, the largest problem
solvable by a single machine will always be constrained by the rate at
which the hardware improves, which has been steadily dwarfed by the
rate at which our data sizes have been increasing over the past decade.
Overall, we think that there are several very strong
reasons to explore the questions of large-scale learning in
cluster environments.

\revised{Previous literature on cluster learning is broad.
Several authors~\citep{Manga95, McDonaldHaMa2010, ZinkevichWeSmLi2010}
have studied approaches that first solve the learning problem independently
on each machine using the portion of the data stored on that machine, and
then average the independent local solutions to obtain the global solution.}
\citet{DuchiAgWa10} propose gossip-style message passing algorithms extending the existing
literature on distributed convex
optimization~\citep{BertsekasTs89}. \citet{LangfordSmZi09} analyze a
delayed version of distributed online learning. \citet{DekelGiShXi10}
consider mini-batch versions of distributed online algorithms which are extended
to delay-based updates in~\citet{AgarwalDu2011}. A recent article of
\citet{BoydPaChPeEc2011} describes an application of the ADMM
technique for distributed learning problems. GraphLab
\citep{LowGoKyBiGuHe10} is a parallel computation framework on graphs.
The closest to our work are optimization approaches based on
centralized algorithms with parallelized gradient
computation~\citep{NashSo89,TeoLeSmVi07}. To our knowledge, all
previous versions of algorithms based on parallelized gradient
computation rely on MPI
implementations.\footnote{\url{http://www.mcs.anl.gov/research/projects/mpi/}}
Finally, the large-scale learning system Sibyl \citep[currently
  unpublished, but see the talks][]{ChandraEtAl10,ChandraEtAl12}
implements a distributed boosting approach. It can be used to solve
the problems of form \eqref{eqn:objective} at the scales similar to
those reported in this paper, but it runs on a proprietary
architecture and many implementation details are missing, so doing a
fair comparison is currently not possible. We attempt to compare the
performance of our algorithm with the published Sibyl performance in
Section~\ref{sec:results}.

All of the aforementioned approaches (perhaps with the exception of
Sibyl) seem to leave something to be desired empirically when deployed
on large clusters.  In particular, their \emph{learning
  throughput}---measured as the input size divided by the wall-clock
running time of the entire learning algorithm---is smaller than the
I/O interface of a single machine for almost all parallel learning
algorithms~\citep[Part III, page 8]{survey}. The I/O interface is an
upper bound on the speed of the fastest \revised{single-machine} algorithm since all
\revised{single-machine} algorithms are limited by the network interface in
acquiring data.  In contrast, we were able to achieve a learning
throughput of 500M features/s, which is about a factor of 5 faster
than the 1Gb/s network interface of any one node. This learning
throughput was achieved on a cluster of 1000 nodes. Each node accessed
its local examples 10 times during the course of the algorithm, so the
per-node processing speeds were 5M features/s.  We discuss our
throughput results in more detail in Section~\ref{sec:results}, and
contrast them with results reported for Sibyl.

Two difficulties bedevil easy parallel machine learning:
\begin{enumerate}
\item Efficient large-scale parallel learning algorithms must occur
  on a data-centric computing platform (such as Hadoop) \revised{to prevent
  data transfer overheads}. These platforms typically do \emph{not} support
  the \revised{full generality} of MPI operations.
\item Existing data-centric platforms often lack efficient mechanisms
  for state synchronization and force both refactoring and rewriting
  of existing learning algorithms.
\end{enumerate}

We effectively deal with both of these issues.  Our system is
compatible with MapReduce clusters such as Hadoop (unlike MPI-based
systems) and minimal additional programming effort is required to
parallelize existing learning algorithms (unlike MapReduce
approaches). In essence, an existing \revised{implementation of
a learning algorithm} need only
insert a few strategic library calls to switch from learning on one
machine to learning on a thousand machines.

One of the key components in our system is a communication
infrastructure that efficiently accumulates and broadcasts values
across all nodes of a computation.  It is functionally similar to MPI
AllReduce (hence we use the name), but it takes advantage of and is
compatible with Hadoop so that programs are easily moved to data,
automatic restarts on failure provide robustness, and speculative
execution speeds up completion.  Our optimization algorithm is a hybrid
online+batch algorithm \revised{with rapid convergence
and only small synchronization overhead, which makes it
a particularly good fit for the distributed environment.}

\revised{In Section~\ref{sec:com} we describe our approach and
our communication infrastructure in more detail. The core
of the paper is Section~\ref{sec:expts} where we
conduct many experiments evaluating our design choices
and comparing our approach with existing algorithms.}
In
Section~\ref{sec:complexity} we \revised{provide some theoretical
  intuition for our design, and contrast
our approach with previous work. We} conclude with a discussion in
Section~\ref{sec:discuss}.

\section{Computation and Communication Framework}
\label{sec:com}

MapReduce~\citep{DeanGh2008} and its open source implementation
Hadoop\footnote{\texttt{http://hadoop.apache.org/}} have become the
overwhelmingly favorite platforms for distributed data processing.
However, the abstraction is rather ill-suited for machine
learning algorithms as several researchers in the field have observed
\citep{LowGoKyBiGuHe10,MatMoTaAnJuMuMiScIo11}, because it does not
easily allow iterative algorithms, such as typical optimization
algorithms used to solve the problem~\eqref{eqn:objective}.

\subsection{Hadoop-compatible AllReduce}

\begin{figure}
	\centerline{\includegraphics[width=0.6\textwidth]{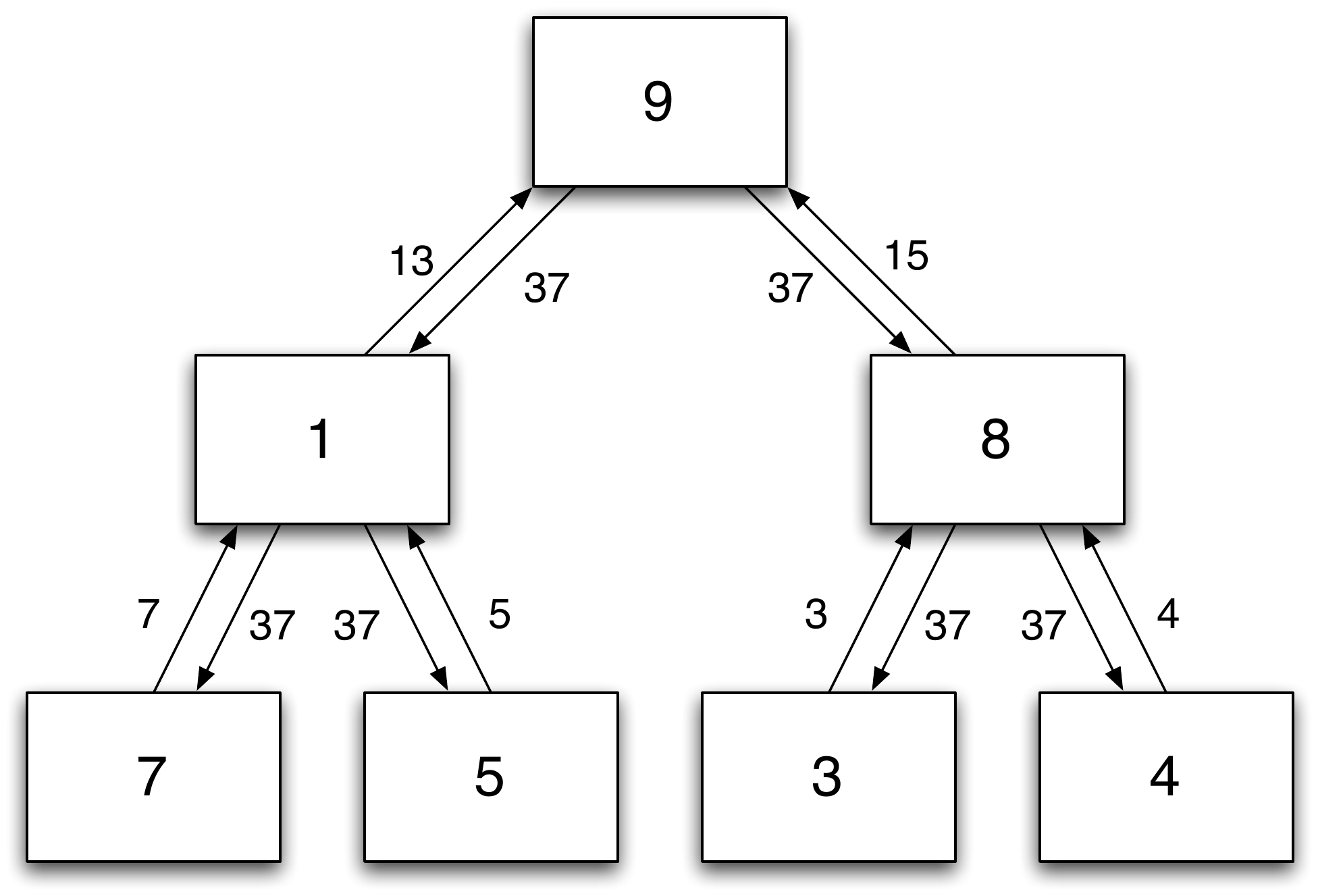}}
	\caption{AllReduce operation. Initially, each node holds its
          own value. Values are passed up the tree and summed, until
          the global sum is obtained in the root node (reduce
          phase). The global sum is then passed back down to all other
          nodes (broadcast phase). At the end, each node contains the
          global sum.}
        \label{fig:allreduce}
\end{figure}

AllReduce is a more suitable abstraction for machine learning
algorithms. AllReduce is an operation where every node starts with a
number and ends up with the sum of the numbers across all the nodes
(hence the name).  A
typical implementation imposes a tree structure on the
communicating nodes and proceeds in two phases:
numbers are first summed up the tree (the
\emph{reduce} phase) and then broadcast down to all the nodes
(the \emph{broadcast} phase), see Figure~\ref{fig:allreduce} for a graphical
illustration. When doing summing or averaging of a long vector, such
as the weight vector $\w$ in the optimization~\eqref{eqn:objective},
the reduce and broadcast operations can be pipelined over the vector
entries and hence the latency of going up and down the tree becomes
negligible on a typical Hadoop cluster. This is the main optimization
we do within the AllReduce architecture.  While other (potentially
more efficient or simpler) architectures for AllReduce are possible,
in our experiments in Section~\ref{sec:expts} we will see that the
time spent in AllReduce operation is negligible compared with the
computation time and stalling time while waiting for other nodes.
Therefore, we do not attempt to optimize the architecture further.

For problems of the form~\eqref{eqn:objective}, AllReduce provides
straightforward parallelization \revised{of gradient-based optimization algorithms such as
gradient descent or L-BFGS---gradients are accumulated locally, and the global
gradient is obtained by AllReduce.}
In
general, any statistical query algorithm~\citep{Kearns93} can be
parallelized with AllReduce with only a handful of additional lines of
code.  This approach also easily implements averaging parameters of
online learning algorithms.

An implementation of AllReduce is available in the MPI
package. However, it is not easy to run MPI on top of existing Hadoop
clusters \citep{YeChChZh09}. Moreover, MPI implements little fault
tolerance, with the bulk of robustness left to the programmer.

To address the reliability issues better, we developed an
implementation of AllReduce that is compatible with Hadoop.
\revised{Our implementation works as follows.} We initialize
a spanning tree server on the gateway node to the Hadoop cluster. We
then launch a map-only (alternatively reduce-only) job where each
mapper processes a subset of the data. Each mapper is supplied with
the IP address of the gateway node, to which it connects as the first
step. Once all the mappers \revised{are} launched and connected to the
spanning tree server, it creates a (nearly balanced) binary tree on
these nodes. Each node is given the IP addresses of its parent and
child nodes in the tree, allowing it to establish TCP connections with
them. \revised{All the nodes are} now ready to pass messages up and
down the tree. The actual communication between the nodes is all
implemented directly using C++ sockets and does not rely on any Hadoop
services.  Implementation of AllReduce using a single tree is clearly
less desirable than MapReduce in terms of reliability, because if any
individual node fails, the entire computation fails. To deal with
this \revised{problem}, we use a simple trick \revised{described} below
which makes AllReduce reliable
enough to use in practice for computations up to 10K node hours.

\revised{It is noteworthy that the idea of using AllReduce for iterative computations
has recently gained traction in the Hadoop community. For instance,
\emph{Knitting Boar}
is an abstraction on top of YARN---the new scheduler for
Hadoop---which is ``similar in nature to the AllReduce primitive''.\footnote{See
\url{https://github.com/jpatanooga/KnittingBoar/wiki/IterativeReduce}.}}

\subsection{Proposed Algorithm}
%\label{sec:algos}

Our main algorithm is a hybrid online+batch approach. \revised{Pure
online and pure batch learning algorithms have some desirable features,
on which we build, and some drawbacks, which we overcome.
For instance,} an attractive feature of online learning algorithms is that
they optimize the objective to a rough precision quite fast, in
just a handful of passes over the data. The inherent sequential nature
of these algorithms, however, makes them tricky to parallelize and we
discuss the drawbacks of some of the attempts at doing so in
Section~\ref{sec:complexity}. Batch learning algorithms such as Newton
and quasi-Newton methods (e.g., L-BFGS), on the other hand, are great
at optimizing the objective to a high accuracy, once they are in a
\emph{good neighborhood} of the optimal solution. But the algorithms
can be quite slow in reaching this good neighborhood. Generalization
of these approaches to distributed setups is rather straightforward,
only requiring aggregation across nodes after every iteration, as has
been noted in previous research~\citep{TeoLeSmVi07}.

We attempt to reap the benefits and avoid the drawbacks of both above
approaches through our hybrid method. We start with each node making
one online pass over its local data according to adaptive gradient
updates~\citep{DuchiHaSi2010,McMahanSt2010} modified for loss
non-linearity~\citep{KarampatziakisLa2011}. We notice that each online
pass happens completely \emph{asynchronously} without any
communication between the nodes, and we can afford to do so since we
are only seeking to get into a good neighborhood of the optimal
solution rather than recovering it to a high precision at this first
stage. AllReduce is used to average these weights non-uniformly
according to locally accumulated gradient squares. Concretely, node
$k$ maintains a local weight vector $\w^k$ and a diagonal matrix
$\G^k$ based on the gradient squares in the adaptive gradient update
rule (see Algorithm \ref{alg:sgd}).  We compute the following weighted
average over all $m$ nodes
\begin{equation}
  \bar{\w} = \left(\sum_{k=1}^m \G^k\right)^{-1}\left(\sum_{k=1}^m
  \G^k \w^k\right).
  \label{eqn:average}
\end{equation}
This has the effect of weighting each dimension according to how
``confident'' each node is in its weight (i.e., more weight is
assigned to a given parameter of a given node if that node has seen
more examples with the corresponding feature).  We note that this
averaging can indeed be implemented using AllReduce by two calls to
the routine since \revised{the matrices} $\G^k$ are only diagonal.

This solution $\bar{\w}$ is used to initialize
L-BFGS~\citep{Nocedal1980} with the standard Jacobi preconditioner,
with the expectation that the online stage gives us a good
\emph{warmstart} for L-BFGS.  At each iteration, global gradients are
  obtained by summing up local gradients via AllReduce, while all the other operations can be done
locally at each node. The algorithm benefits from the fast initial
reduction of error provided by an online algorithm, and rapid
convergence in a good neighborhood guaranteed by quasi-Newton
algorithms. We again point out that the number of communication
operations is relatively small throughout this process.

\revised{In addition to hybrid strategy, we also evaluate}
repeated online learning with
averaging using the adaptive updates. In this setting, each node
performs an online pass over its data and then we average the weights
according to Equation~\ref{eqn:average}. We average the scaling
matrices similarly
\[
  \bar{\G} = \left(\sum_{k=1}^m \G^k\right)^{-1}\left(\sum_{k=1}^m
  (\G^k)^2\right)
  \label{eqn:average:G}
\]
and use this averaged state to start a new online pass over the
data. This strategy is similar to those proposed
by~\citet{McDonaldHaMa2010} and~\citet{HallGiMa2010} for different
online learning algorithms. We will see in the next section that this
strategy can be very effective at getting a moderately small test
error very fast, \revised{but its convergence slows down and it might be
too slow at reaching the optimal test error}.

All strategies described above share the same processing
structure. They carry out several iterations, each of which can be
broken into three phases: (1) Pass through the entire local portion of
the dataset and accumulate the result as a vector of size~$d$ (i.e.,
the same size as the parameter vector).  (2) Carry out AllReduce
operation on a vector of size~$d$.  (3) Do some additional processing
and updating of the parameter vector.

The key point to notice is that in typical applications the local
dataset will be orders of magnitude larger than the parameter vector,
hence the communication after each pass is much more compact than
transmitting the entire local dataset. The second point is that each
iteration is naturally a MapReduce operation. The main reason that we
expect to benefit by AllReduce is because of the iterative nature of
these algorithms and the shared state between iterations.

Our implementation is available as part of the open source project
Vowpal Wabbit~\citep{LangfordLiSt07} and is summarized in
Algorithm~\ref{alg:learn}.  It makes use of stochastic gradient
descent (Algorithm~\ref{alg:sgd}) for the initial pass.

\begin{algorithm}
	\caption{Stochastic gradient descent algorithm on a single
          node using adaptive gradient update
          \citep{DuchiHaSi2010,McMahanSt2010}.}
	\label{alg:sgd}\small
	\begin{algorithmic}
		\REQUIRE Invariance update function $s$
                \STATE    ~~~~~~~~~~~(see \citealp{KarampatziakisLa2011})
		\STATE $\w=\boldsymbol{0},\,\G=\mathbf{I}$
		\FORALL{$(\x,y)$ in training set}
			\STATE $\g \gets \nabla_{\!\w}\,\ell(\w^\tr\x;\, y)$
			\STATE $\displaystyle\w \gets \w - s(\w,\x,y)\G^{-1/2}\g$
			\STATE $G_{jj} \gets G_{jj} + g_j^2$ for all $j=1,\dots,d$
		\ENDFOR
	\end{algorithmic}
\end{algorithm}

\begin{algorithm}
	\caption{Sketch of the proposed learning architecture}
	\label{alg:learn}\small
	\begin{algorithmic}
		\REQUIRE Data split across nodes
		\FORALL{nodes $k$}
		    \STATE $\w^k$ = result of stochastic gradient descent
                   on the data of node $k$ using Algorithm \ref{alg:sgd}.
		\STATE Compute the weighted average $\bar{\w}$ as in \eqref{eqn:average}
               using AllReduce.
		\STATE Start a preconditioned L-BFGS optimization from $\bar{\w}$.
			\FOR{$t=1,\dots,T$}
				\STATE Compute $\g^k$ the (local batch) gradient of examples on node $k$.
			    \STATE Compute $\g = \sum_{k=1}^m \g^k$ using AllReduce.
			    \STATE Add the regularization part in the gradient.
				\STATE Take an L-BFGS step.
			\ENDFOR
		\ENDFOR
	\end{algorithmic}
\end{algorithm}

\subsection{Speculative Execution}

Large clusters of machines are typically busy with many jobs
which use the cluster unevenly, resulting in one of a thousand
nodes being very slow. To avoid this, Hadoop can speculatively execute
a job on identical data, using the first job to finish and killing the
other one. In our framework, it can be tricky to handle duplicates
once a spanning tree topology is created for AllReduce. For this
reason, we delay the initialization of the spanning tree until each
node completes the first pass over the data, building the spanning tree on
only the speculative execution survivors.  The net effect of this
speculative execution trick is perhaps another order of magnitude of
scalability and reliability in practice. Indeed, we found the system
reliable enough for up to 1000 nodes running failure-free for hundreds
of trials (of typical length up to 2 hours).  This \revised{level
of fault tolerance} highlights the benefits of a
Hadoop-compatible implementation of AllReduce. We will show the
substantial gains from speculative execution in mitigating the ``slow
node'' problem in the experiments.

\section{Experiments}
\label{sec:expts}

\subsection{Datasets}
\label{sec:datasets}

\paragraph{Display Advertising}

In online advertising, given a user visiting a publisher page, the
problem is to select the best advertisement for that user. A key
element in this matching problem is the click-through rate (CTR)
estimation: what is the probability that a given ad will be clicked
\revised{on,} given some context (user, page visited)? Indeed, in a
cost-per-click (CPC) campaign, the advertiser only pays when the ad
gets clicked, so even modest improvements in predictive accuracy
directly affect revenue.

Training data contains user visits, which either resulted in a click
on the ad (positive examples with $y_i=1$), or did not result in a
click (negative examples with $y_i=0$). We estimate the click
probabilities by logistic regression with $L_2$ regularization.  The
regularization coefficient is chosen from a small set to obtain the
best test performance. The user visit is represented by binary
indicator features encoding the user, page, ad, as well as
conjunctions of these features. Some of the features include
identifiers of the ad, advertiser, publisher and visited page. These
features are hashed \citep{WeinbergerDaLaSmAt2009} and each training
example ends up being represented as a sparse binary vector of
dimension~$2^{24}$ with around 125 non-zero elements. Let us
illustrate the construction of a conjunction feature with an
example. Imagine that an ad from \texttt{etrade} was placed on
\texttt{finance.yahoo.com}. Let $h$ be a 24 bit hash of the string
``\texttt{publisher=finance.yahoo.com and advertiser=etrade}''.  Then
the (publisher, advertiser) conjunction is encoded by setting to 1 the
$h$-th entry of the feature vector for that example.

Since the data is unbalanced (low CTR) and because of the large number of
examples, we subsample the negative examples resulting in a class
ratio of about 2 negatives for 1 positive, and use a large test set
drawn from days later than the training set.  There are 2.3B examples
in the training set. \revised{More characteristics of this dataset and
  modeling details can be found in~\citep{ChaManRos13}}

\paragraph{Splice Site Recognition}

The problem consists of recognizing a human acceptor splice site
\citep{SonnenburgFr10}. We consider this learning task because this
is, as far as \revised{we} know, the largest public dataset for which
subsampling is not an effective learning strategy. \citet{SonRaeRie07}
introduced the {\em weighted degree kernel} to learn over DNA
sequences. They also proposed an SVM training algorithm for that
kernel for which learning over 10M sequences took 24 days.
\revised{\citet{SonnenburgFr10}} proposed an improved training algorithm,
in which the weight vector---in the feature space induced by
the kernel---is learned, but the feature vectors are never explicitly
computed. This resulted in a faster training: 3 days with 50M
sequences.

We solve this problem by $L_2$-regularized logistic regression. Again,
the regularization coefficient is chosen from a small set to optimize
test set performance.  We follow the same experimental protocol as in
\citet{SonnenburgFr10}: we use the same training and test sets of
respectively 50M and 4.6M samples. We also consider the same kernel of
degree $d=20$ and hash size $\gamma=12$. The feature space induced by
this kernel has dimensionality 11,725,480. The number of non-zero
features per sequence is about 3,300. Unlike \citet{SonnenburgFr10},
we explicitly compute the feature space representation of the examples,
yielding about 3TB of data.  This explicit representation is a
\emph{disadvantage} we impose on our method to simplify implementation.

\subsection{Results}
\label{sec:results}

\paragraph{Effect of Subsampling}

The easiest way to deal with a very large training set is to reduce
it by subsampling as discussed in the introduction. Sometimes similar test errors can
be achieved with smaller training sets and there is no need for large-scale
learning.  For splice site recognition, Table 2
of \citet{SonnenburgFr10} shows that smaller training sets do hurt the
area under the precision/recall curve on the test set.

For display advertising, we subsampled the data at 1\% and 10\%. The
results in Table~\ref{tbl:learningcurve} show that there is a
noticeable drop in accuracy after subsampling. Note that even if the
drop does not appear large at a first sight, it can cause a
substantial loss of revenue. Thus, for both datasets, the entire
training dataset is needed to achieve optimal performances.

The three metrics reported in Table \ref{tbl:learningcurve} are area
under the ROC curve (auROC), area under the precision/recall curve
(auPRC) and negative log-likelihood (NLL). Since auPRC is the most
sensitive metric, we report test results using that metric in the rest
of the paper. This is also the metric used in \citet{SonnenburgFr10}.

\begin{table}
\caption{Test performance on the display advertising problem as a
  function of the subsampling rate, according to three metrics: area
  under ROC curve (auROC), area under precision/recall curve (auPRC),
  and negative log likelihood (NLL).}%
% Note: Regularization=300. 20 iterations.
\label{tbl:learningcurve}%
\centering\small%
	  \begin{tabular}{l|ccc}
		& 1\% & 10\% & 100\% \\ \hline
		auROC & 0.8178 & 0.8301 & 0.8344 \\
		auPRC & 0.4505 & 0.4753 & 0.4856 \\
		NLL & 0.2654 & 0.2582 & 0.2554
	\end{tabular}
\end{table}

\paragraph{Running Time}
We ran 5 iterations of L-BFGS on the splice site data with 1000
nodes. On each node, we recorded for every iteration the time spent in
AllReduce and the computing time---defined as the time not spent in
AllReduce.  The time spent in AllReduce can further be divided into
stall time---waiting for the other nodes to finish their
computation---and communication time. The communication time can
be estimated by
taking the minimum value of the AllReduce times across nodes.

The distribution of the computing times is of particular interest
because the speed of our algorithm depends on the slowest
node. Statistics are shown in Table \ref{tbl:times}. It appears that
computing times are concentrated around the median, but
there are a few outliers. Without speculative execution, one single
node was about 10 times slower than the other nodes; this has the
catastrophic consequence of slowing down the entire process by a
factor 10. The table shows that
the use of speculative execution successfully mitigates this issue.

\begin{table}
	\caption{Distribution of computing times (in seconds) over 1000
          nodes \revised{with and without speculative execution}.
          First three columns are quantiles. Times are average
          per iteration (excluding the first one) for the splice site
          recognition problem.}%
	\label{tbl:times}%
\centering\small%
	\begin{tabular}{l|cccc|c}
		& 5\% & 50\% & 95\% & Max & Comm. time\\ \hline
		Without spec.~execution & 29 & 34 & 60 & 758 & 26 \\
		With spec.~execution & 29 & 33 & 49 & 63 & 10 \\
	\end{tabular}
\end{table}

\begin{figure}
\begin{center}
\includegraphics[width=0.6\textwidth]{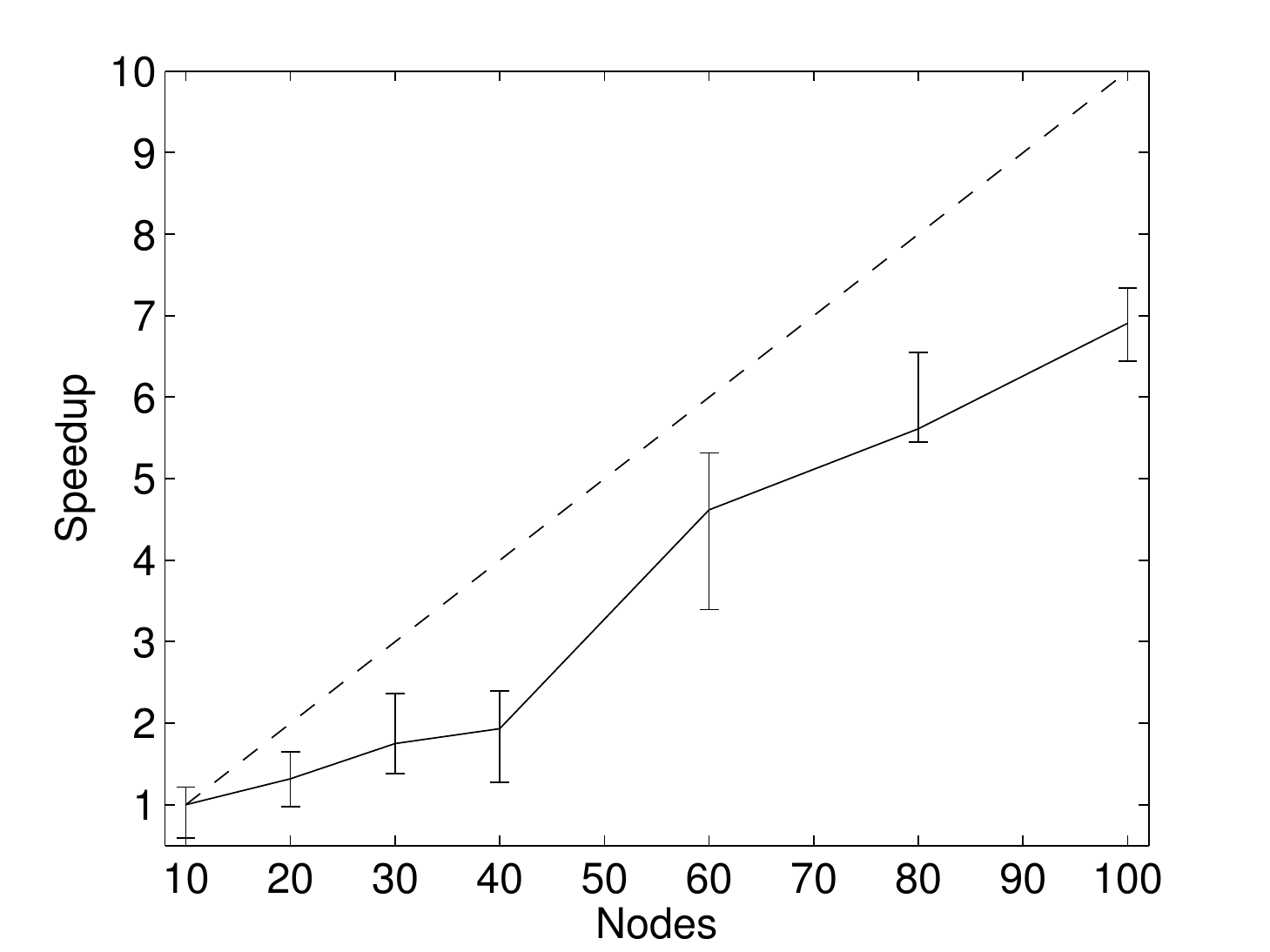}
\end{center}
\caption{Speed-up for obtaining a fixed test error, on the display
  advertising problem, relative to the run with 10 nodes, as a
  function of the number of nodes. The dashed line corresponds to the ideal
  speed-up, the solid line is the average speed-up over 10 repetitions,
  and the bars indicate maximum and minimum values.}
\label{fig:speedup}
\end{figure}

We now study the running time as a function of the number of
nodes. For the display advertising problem, we varied the number of
nodes from 10 to 100 and computed the speed-up factor relative to the
run with 10 nodes. In each case, we measured the amount of time needed
to get to a fixed test error. Since there can be significant
variations from one run to the other---mostly because of the cluster
utilization---each run was repeated 10 times. Results are reported in
Figure \ref{fig:speedup}. We note that speculative execution was not
turned on in this experiment, and we expect better speed-ups with
speculative execution. In particular, we expect that the main reason
for the departure from the ideal speed-up curve is the ``slow node''
problem (as opposed to the aspects of the AllReduce communication
implementation), which is highlighted also in the next experiment.

\begin{table}
	\caption{Computing times to obtain a fixed test error on the splice site
          recognition data, using different numbers of nodes.
          The first 3 rows are averages per iteration (excluding
          the first pass over the data).}%
	\label{tbl:speedup}%
\centering\small%
	\begin{tabular}{l|cccc}
		Nodes & 100 & 200 & 500 & 1000 \\ \hline
		Comm time / pass & 5 & 12 & 9 & 16 \\
		Median comp time / pass & 167 & 105 & 43 & 34 \\
		Max comp time / pass & 462 & 271 & 172 & 95 \\ \hline
		Wall clock time & 3677 & 2120 & 938 & 813
	\end{tabular}
\end{table}

\begin{figure}
\begin{center}
\includegraphics[width=0.6\textwidth]{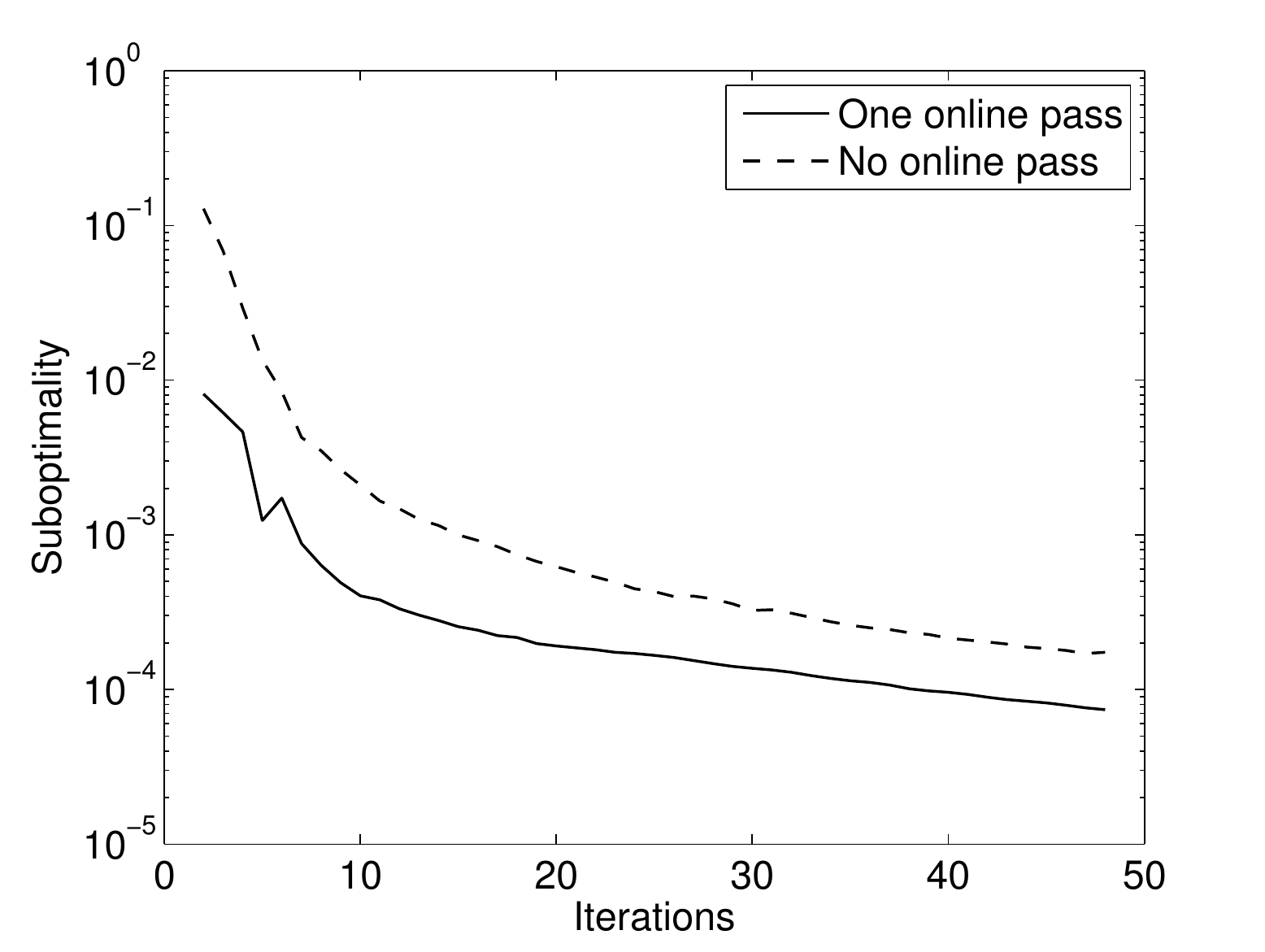}
\caption{Effect of initializing the L-BFGS optimization by an average
  solution from online runs on individual nodes. Suboptimality is the
  difference between the objective value on the training data and the
  optimal value obtained by running the algorithm to convergence.}
\label{fig:warmstart}
\end{center}
\end{figure}

Table \ref{tbl:speedup} shows the running times for attaining a fixed
test error as a function of the number of nodes on the splice site
recognition problem. Unlike Figure \ref{fig:speedup}, these timing
experiments have not been repeated and thus there is a relatively large
uncertainty in their expected values. It can be seen from Tables
\ref{tbl:times} and \ref{tbl:speedup} that even with as many as 1000
nodes, communication is not the bottleneck. One of the main challenges
instead is the ``slow node" issue. This is mitigated to some degree by
speculative execution, but as the number of nodes increases, so does
the likelihood of hitting slow nodes.

\paragraph{Large Experiment and Comparison with Sibyl}
We also experimented with an 8 times larger version of the display
advertising data (16B examples). Using 1000 nodes and 10 passes over
the data, the training took only 70 minutes.\footnote{As mentioned
  before, there can be substantial variations in timing between
  different runs; this one was done when the cluster was not
  \revised{too} occupied.}  Since each example is described by 125
non-zero features, the average processing speed was
\[
   16\text{B}\times 10 \times 125\text{ features} / 1000 \text{ nodes} / 70 \text{ minutes}
   \;
   =
   \;
   4.7\text{ M features/node/s}
\enspace.
\]
The overall learning throughput was
\[
   16\text{B}\times 125\text{ features} / 70 \text{ minutes}
   \;
   =
   \;
   470\text{ M features/s}
\enspace.
\]
We briefly compare this with a result reported for the distributed
boosting system Sibyl for a run on
970 cores~\citep[slide 24]{ChandraEtAl12}. The run was done over
129.1B examples, with 54.61 non-zero features per example. The
reported processing speed was 2.3M features/core/s (which is a factor
of two slower than our achieved processing speed).  The reported
number of iterations was 10--50, which would lead to the final learning
throughput in the range 45--223 M features/s, i.e., the result appears
to be slower by a factor of 2--10.

%Note that there are many uncontrolled quantities in any comparison due
%to Sibyl not being published. We expect that, if anything, our
%approach is at a \emph{disadvantage} compared with Sibyl: our computing nodes are low-end servers,
%the data resting state is an example-per-line pre-dictionary text format,
%Hadoop is by all reports rather inferior to Google's computing architecture,
%and all of our timings were done on a multi-user
%cluster which we had little control over. With other choices, which
%were perhaps made for Sibyl, we would expect the algorithmic
%advantage of our approach to be even larger.}

\paragraph{Online and Batch Learning}
We now investigate the speed of convergence of three different learning
 strategies: batch, online and hybrid. We are interested in how fast the algorithms minimize the training objective as well as the
 test error.

% \begin{table}
%       \caption{auPRC for two online algorithms on the display advertising problem.}
%         \label{tbl:online}
%        \begin{center}
%         \begin{tabular}{l|cc}
%                 & Iter 1 & Iter 20 \\ \hline
%                 VW-adaptive & 0.4685 & 0.4846 \\
%                 SGD & 0.3158 & 0.3831
%         \end{tabular}
%     \end{center}
%  \end{table}
%
% First Table \ref{tbl:online} clearly shows that VWs optimized
% adaptive online learning algorithm converges much faster than a
% standard gradient descent rule. In the rest of this section, we thus
% only consider the optimized online learning algorithm.

Figure \ref{fig:warmstart} compares how fast the two
learning strategies---batch with and without an initial online pass---
optimize the training objective. It plots the optimality gap, defined as the difference
between the current objective function and the optimal one
(i.e., the minimum value of the objective~\eqref{eqn:objective}), as a
function of the number iterations. From this figure we can see that
the initial online pass results in a saving of about 10--15 iterations.

\begin{figure*}[tbp]
\small
\hspace{0.1in}
\hfill\textsf{Splice site recognition}
\hfill
\hfill\textsf{Display advertising}\hfill\hphantom{}\hspace{0.05in}\\%
\includegraphics[width=\textwidth]{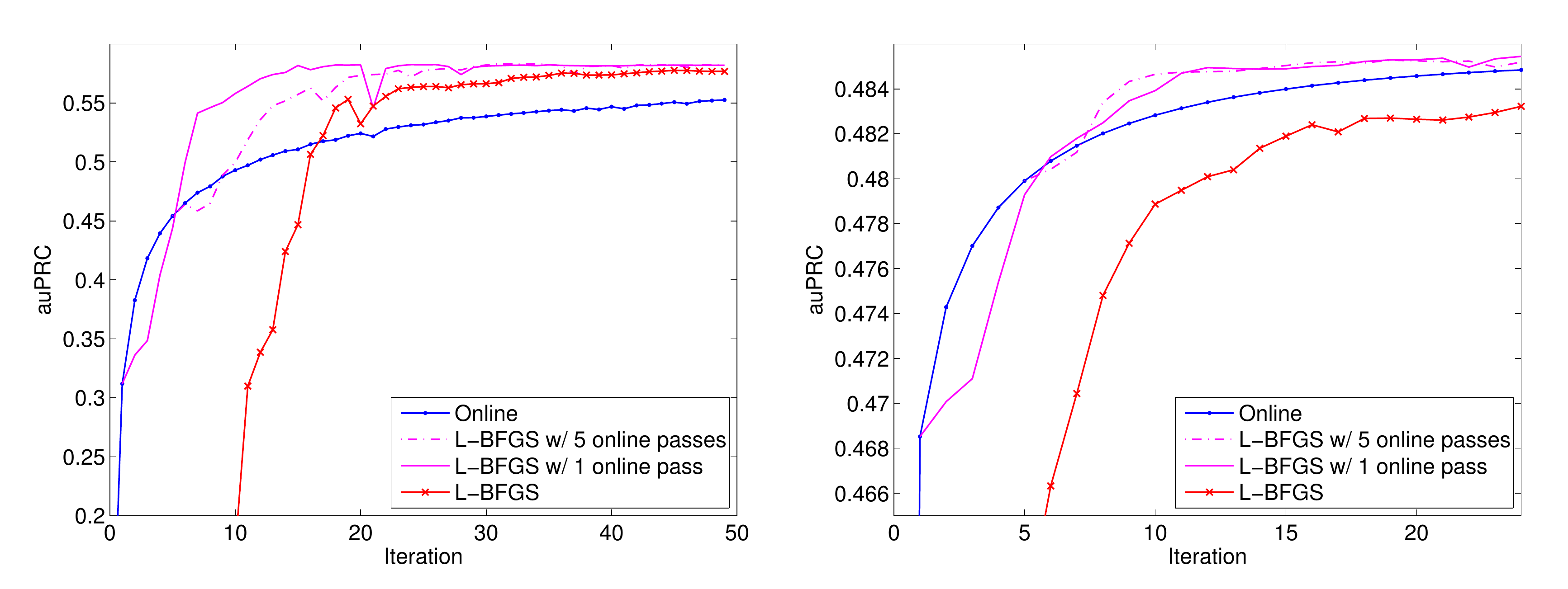}
\caption{Test auPRC for 4 different learning strategies. \revised{Note
    that the online and hybrid curves overlap during the warmstart
    phase (of either 1 or 5 online passes).}}
\label{fig:testerror}
\end{figure*}

Figure \ref{fig:testerror} shows the test auPRC on both datasets as
a function of the number of iterations for 4 different strategies:
only online learning, only L-BFGS learning, and 2 hybrid methods
consisting of 1 or 5 passes of online learning followed by L-BFGS
optimization. L-BFGS with one online pass appears to be the most
effective strategy.

For the splice site recognition problem, an initial online pass and 14
L-BFGS iterations yield an auPRC of 0.581, which is just a bit higher
than results of~\citet{SonnenburgFr10}. This was achieved in 1960 seconds using
500 machines, resulting in a 68 speed-up factor (132,581 seconds on a
single machine reported in Table 2 of \citet{SonnenburgFr10}).  This
seems rather poor compared with the ideal 500 speed-up factor, but
recall that we used explicit feature representation which creates a
significant overhead.

%Note LL online = 0.345797; BFGS = 0.337665 (on training set)

%\paragraph{Averaging}

%\begin{table}
%\caption{auPRC after one online pass followed by 5 L-BFGS iterations.}
%\label{tab:average}
%\begin{center}
%\begin{tabular}{l|ccc}
% & No avg. & Unif. avg. & Weighted avg. \\ \hline
%Display & 0.4729 & 0.4815 & 0.4810 \\
%Splice & 0.4188 & 0.3164 & 0.4996
%\end{tabular}
%\end{center}
%\end{table}

%Next, we investigate the effect of the averaging strategy at the
%end of an online pass.
%Table~\ref{tab:average} compares picking one online run at random,
%using uniform weight averaging, or using non-uniform weight averaging
%according to Equation~\ref{eqn:average} from adaptive updates.  Note
%that the random pick for the splice site recognition dataset
%was apparently lucky, and that
%weighted averaging works consistently well.

\subsection{Comparison with Previous Approaches}

\paragraph{AllReduce vs. MapReduce}
The standard way of using MapReduce for iterative machine learning
algorithms is the following \citep{ChuKiLiYuBrNgOl07}: every iteration
is a MapReduce job where the mappers compute some local statistics (such as
a gradient) and the reducers sum them up. This is ineffective because
each iteration has large overheads (job scheduling, data transfer,
data parsing, etc.).  We have an internal implementation of such a MapReduce
algorithm. We updated this code to use AllReduce instead and compared
both versions of the code in Table \ref{tbl:mapreduce}. This table
confirms that Hadoop MapReduce has substantial overheads since the
training time is not much affected by the dataset size.  The speed-up
factor of AllReduce over Hadoop's MapReduce can become extremely large
for smaller datasets, and remains noticeable even for the largest
datasets.
It is also noteworthy that {\em all} algorithms described in
\citet{ChuKiLiYuBrNgOl07} can be parallelized with AllReduce, plus
further algorithms such as parameter averaging approaches.

\begin{table}
	\caption{Average training time per iteration of an internal
          logistic regression implementation using either MapReduce or
          AllReduce for gradients aggregation. The dataset is the
          display advertising one and a subset of it.}%
	\label{tbl:mapreduce}%
\centering\small%
	\begin{tabular}{l|cc}
		& Full size & 10\% sample \\ \hline
		MapReduce & 1690 & 1322 \\
		AllReduce & 670 & 59
	\end{tabular}
\end{table}
	
\paragraph{Overcomplete Average}

We implemented oversampled stochastic gradient with final
averaging~\citep{ZinkevichWeSmLi2010}, and compared its performance to
our algorithm.  We used stochastic gradient descent with the learning
rate in the $t$-th iteration as
\[
  \eta_t = \frac{1}{L+\gamma\sqrt{t}}
\enspace.
\]
We tuned $\gamma$ and $L$ on a small subset of the dataset.

\begin{figure}
\small
\hspace{0.1in}
\hfill\textsf{Splice site recognition}
\hfill
\hfill\textsf{Display advertising}\hfill\hphantom{}\hspace{0.05in}\\%
\includegraphics[width=0.5\textwidth]{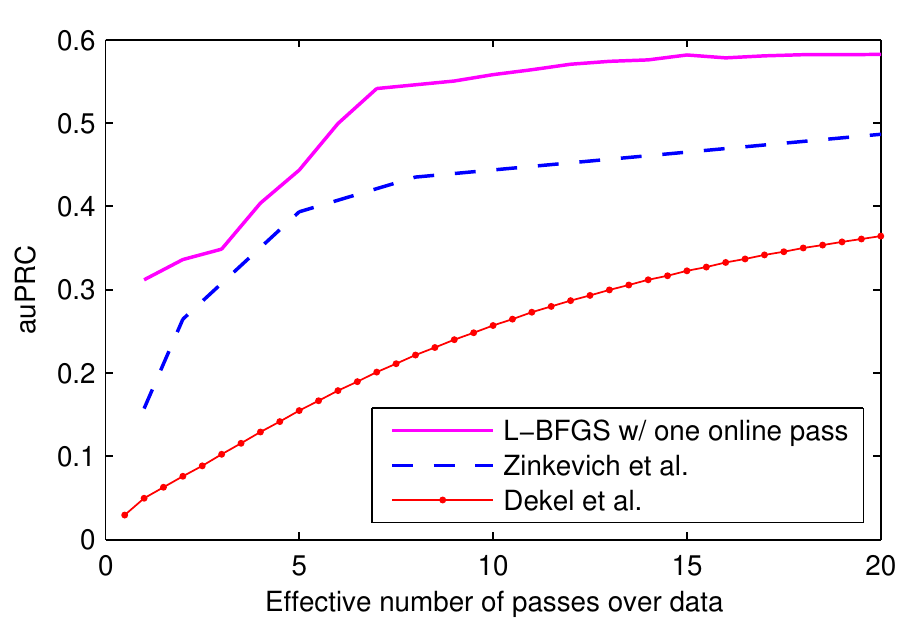}
\includegraphics[width=0.5\textwidth]{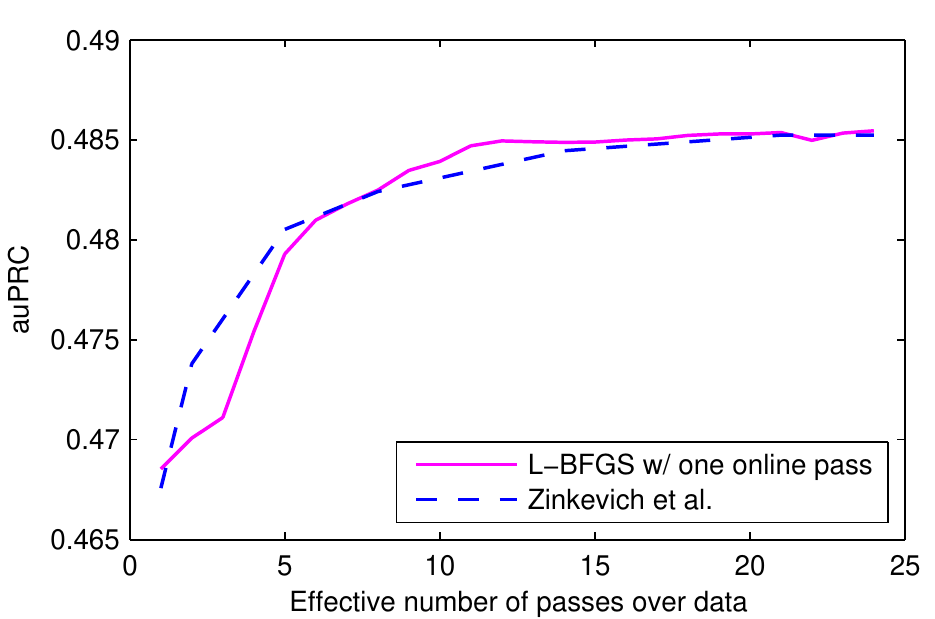}
\caption{Test auPRC for different learning strategies as a function of
  the effective number of passes over data. In L-BFGS, it corresponds to
  iterations of the optimization. In
  overcomplete SGD with averaging (Zinkevich et al.), it corresponds to
  the replication coefficient.}
\label{fig:marty}
\end{figure}

In Figure~\ref{fig:marty}, we see that the oversampled SGD is
competitive with our approach on the display advertising data,
but its convergence is much slower on splice site data.

\paragraph{Parallel Online Mini-batch}

\citet{DekelGiShXi10} propose to perform online convex optimization
using stochastic gradients accumulated in small mini-batches across
all nodes. We implemented a version of their algorithm using
AllReduce. They suggest global minibatch sizes of no more than
$b\propto \sqrt{n}$.  On $m$ nodes, each node accumulates gradients
  from $b/m$ examples, then an AllReduce operation is carried out,
  yielding the mini-batch gradient, and each node performs a
  stochastic gradient update with the learning rate of the form
\[
     \eta_t = \frac{1}{L+\gamma\sqrt{t/m}}
\enspace.
\]
We tuned $L$ and $\gamma$ on a smaller dataset. In
Figure~\ref{fig:marty}, we report the results on splice site dataset,
using 500 nodes, and mini-batch size $b=100\text{k}$. Twenty passes over the
data thus corresponded to 10k updates.  Due to the overwhelming
communication overhead associated with the updates, the overall
running time was 40 hours. In contrast, L-BFGS took less than an hour
to finish 20 passes while obtaining much superior performance. The
difference in the running time between 1h and 40h is solely due to
communication. Thus, in this instance, we can conservatively conclude
that the communication overhead of 10k mini-batch updates is 39 hours.

We should point out that it is definitely possible that the
mini-batched SGD would reach similar accuracy with much smaller
mini-batch sizes (for 10k updates theory suggests we should use
mini-batch sizes of at most 10k), however, the 39 hour communication
overhead would remain.  Using larger mini-batches, we do expect that
the time to reach 20 passes over data would be smaller (roughly
proportional to the number of mini-batch updates), but according to
theory (as well as our preliminary experiments on smaller subsets of
splice site data), we would have inferior accuracy. Because of the
prohibitive running time, we were not able to tune and evaluate this
algorithm on display advertising dataset.

%% \paragraph{Parallel Online Learning}

%% Finally we compared our approach using the online parallel learning
%% algorithm of \cite{HsuKaLaSm11} using the same online advertising
%% dataset in their paper. We note that this is a substantially smaller
%% dataset with about 10M examples, and 125G non-zero features in the
%% data matrix. We did not run this comparison on our larger datasets
%% since the methods in~\citet{HsuKaLaSm11} do not scale well to a large
%% number of nodes, as evident from Figure 5 of their paper: with 8
%% nodes, the speed-up is only a factor of 2. For both algorithms, we set
%% the number of passes over the data to reach a certain test error. This
%% number turned out to be 18 for the parallel online learning and 20 for
%% our algorithms.  The running time using 8 nodes was 35 minutes for the
%% parallel online learning and 16 minutes for ours.

\section{Communication and Computation Complexity}
\label{sec:complexity}

The two key performance characteristics of any distributed algorithm
are its communication and computation complexity. The aim of this
section is to discuss the complexity of our approach and to compare
it with previous solutions. We hope to clarify the reasons underlying our design
choices and explain the scalability of our system.
We start with a discussion of computational
considerations.

\subsection{Computational Complexity of the Hybrid Approach}
\label{sec:compcomplexity}

In this section, we explain the convergence properties of the hybrid
approach and compare it with other optimization strategies. In order
to have a clean discussion, we make some simplifying assumptions.
We consider the case of only one online pass at each
node. Furthermore, we restrict ourselves to the case of uniform
averaging of weights. Similar analysis does extend to the
non-uniform weighting scheme that we use, but the details are
technical and provide no additional intuition. Before we
embark on any details, it should be clear that the hybrid approach is
always convergent, owing to the convergence of L-BFGS. All the online
learning step initially does is to provide a good warmstart to
L-BFGS. This section aims to provide theoretical intuition
why the gains of such a warmstart can be substantial in certain
problem regimes.

Let $\tell(\w;\,\x,\,y)=\ell(\w^\tr\x;\,y) + \lambda R(\w)/n$ be
the regularized loss function.
We analyze a scaled version of the objective~\eqref{eqn:objective}:
\[
  f(\w) = \frac{1}{n}\, \sum_{i=1}^n \ell(\w^\tr\x_i;\,y_i) + \frac{\lambda}{n}R(\w)
        = \frac{1}{n}\,\sum_{i=1}^n\tell(\w;\,\x_i,\,y_i)
\enspace.
\]
We assume that the cluster is comprised of $m$ nodes, with a total of $n$
data examples distributed uniformly at random across these
nodes. Let us
denote the local objective function at each node as $f^k$:
\begin{equation*}
  f^k(\w) = \frac{m}{n}\, \sum_{i \in S_k} \tell(\w;\,\x_i,\,y_i)
\end{equation*}
where $S_k$ is the set of $n/m$ examples at node $k$. Note that
the global objective $f = (\sum_{k=1}^m f^k)/m$ is the average of the
local objectives. We observe that owing to our random data split, we
are guaranteed that
\begin{equation*}
\E[f^k(\w)] = \E[f(\w)] = \Exy\BigBracks{\tell(\w;\,\x,\,y)}
\end{equation*}
for each $k$, where the expectation is taken over the
distribution from which our examples are drawn. In order to discuss
the convergence properties, we need to make a couple of standard
assumptions regarding the functions $f^k$. First, we assume that the
functions $f^k$ are differentiable, with Lipschitz-continuous
gradients. We also assume that each $f^k$ is strongly convex, at least
in a local neighborhood around the optimum. We note that these
assumptions are unavoidable for the convergence of quasi-Newton
methods such as L-BFGS.

To understand how many passes
over the data are needed for the hybrid approach to minimize $f$
to a precision $\epsilon$, we first analyze the online learning pass at
each node. In this pass, we compute a weight vector $\w^k$ by performing $n/m$ steps
of stochastic gradient descent or some variant
thereof~\citep{DuchiHaSi2010,KarampatziakisLa2011}. Since we performed
only one pass at each node, the resulting $\w^k$ at each node
approximately minimizes $\E[f^k] = \Exy[\tell]$ to a precision $\epsilon^k$
(for the methods we use, we expect $\epsilon^k =
\order(\sqrt{m/n})$). Let us now denote the uniform average $\bar{\w} =
\sum_{k=1}^m \w^k/m$. For this approach, a direct application of
Jensen's inequality yields
\begin{align}
\label{eqn:Jensen}
  \Exy\BigBracks{\tell(\bar{\w};\,\x,y)}
  &
   =
   \Exy\left[ \tell\left( \frac{\sum_{i=1}^m \w^k}{m} ;\,\x,y\right) \right]
   \leq
   \frac{1}{m} \sum_{k=1}^m \Exy\BigBracks{\tell(\w^k;\,\x,y)}
  \\
\notag
  &
   \leq
   \min_\w \Exy\BigBracks{\tell(\w;\,\x,y)} + \frac{1}{m} \sum_{k=1}^m \epsilon^k
   =
   \min_\w \Exy\BigBracks{\tell(\w;\,\x,y)} + \order\left(\sqrt{\frac{m}{n}}\right)
\enspace.
\end{align}
Furthermore, standard sample complexity
arguments~\citep[see, e.g.,][]{BartlettMe02, DevroyeGyLu96}
allow us to bound the function value $f(\w)$ for an arbitrary $\w$ as
\begin{equation}
\label{eqn:devbound}
  \left\lvert f(\w) - \Exy\BigBracks{\tell(\w;\,\x,\,y)}\right\rvert \le \order\left( \frac{1}{\sqrt{n}}
  \right)
\enspace.
\end{equation}
Let $\w^*$ be the minimizer of the training loss function $f$. Then
we can combine the above inequalities as
\begin{align*}
  f(\bar{\w})&
    \le\Exy\BigBracks{
         \tell(\bar{\w};\,\x,\,y)}
        +\order(1/\sqrt{n})
  \\&
    \le\min_\w\Exy\BigBracks{\tell(\w;\,\x,\,y)}+\order(\sqrt{m/n})
  \\&
    \le\Exy\BigBracks{\tell(\w^*;\,\x,\,y)}+\order(\sqrt{m/n})
  \\&
    \le f(\w^*)+\order(\sqrt{m/n})
\end{align*}
where the first inequality follows by \eqref{eqn:devbound}, the second by \eqref{eqn:Jensen},
and the fourth by \eqref{eqn:devbound}. For the remainder of the discussion, we
denote the overall suboptimality of $\bar{\w}$ relative to $\w^*$ by
$\epsilon_0= \order(\sqrt{m/n})$.

Switching over to the L-BFGS phase, we assume that
we are in the linear convergence regime of
L-BFGS~\citep{LiuNo1989}. We denote the contraction factor by $\kappa$,
so that the number of additional L-BFGS passes over data needed to
minimize $f$ to a precision $\epsilon$ is at most
\begin{equation*}
  \kappa \log \frac{\epsilon_0}{\epsilon}
\enspace.
\end{equation*}
Compared to initializing L-BFGS without any warmstart, our hybrid
strategy amounts to overall savings of
\begin{equation*}
  \kappa\log\frac{1}{\epsilon} - \left(1+\kappa \log \frac{\epsilon_0}{\epsilon}\right)
  = \kappa\log\frac{1}{\epsilon_0} -1 = \order\left( \frac{\kappa}{2}
  \log\frac{n}{m}\right) - 1
\end{equation*}
passes over data. In typical applications, we expect $n \gg m$ to
ensure that computation amortize the cost of communication.  As a result,
the improvement due to the warmstart can be quite substantial just
like we observed in our experiments. Furthermore, this part of our
argument is in no way specific to the use of L-BFGS as the batch
optimization algorithm. Similar reasoning holds for any reasonable
(quasi)-Newton method.

We could also consider the alternative choice of just using parallel
online learning without ever switching to a batch optimization
method. The theoretical results in this area, however, are relatively
harder to compare with the hybrid approach. For general online
learning algorithms, previous works study just one local pass of
online learning followed by averaging~\citep{McDonaldHaMa2010}, which
typically cannot guarantee an error smaller than $\epsilon_0$ in our
earlier notation. The repeated averaging approach, discussed and
analyze for the specific case of perceptron algorithm in earlier
work~\citep{McDonaldHaMa2010}, works well in our experiments on the
computational advertising task but does not have easily available
convergence rates beyond the special case of separable data and the 
perceptron algorithm. Nevertheless, one appeal of the hybrid approach
is that it is guaranteed to be competitive with such online
approaches, by the mere virtue of the first online phase.

Overall, we see that the hybrid approach will generally be competitive
with purely online or batch approaches in terms of the computational
complexity. As a final point, we discuss two extreme regimes
where it can and does offer substantial gains. The first regime is
when the data has a significant noise level. In such a scenario, the
level $\epsilon$ of optimization accuracy desired is typically not too large
(intuitive statistical arguments show no reduction in generalization error
for $\epsilon \ll 1/n$). Setting $\epsilon = 1/n$ for a clean comparison, we observe
that the total number of passes for the hybrid method is at most
\begin{equation*}
  1 + \frac{\kappa}{2}(\log(m) + \log(n)),
\end{equation*}
as opposed to $\kappa \log(n)$ for just pure batch optimization. When
$m \ll n$, this shows that the online warmstart can cut down the
number of passes almost by a factor of 2. We do note that in such high
noise regimes, pure online approaches can often succeed, as we observed
with our advertising data.

The second extreme is when our data is essentially noiseless, so that
the desired accuracy $\epsilon$ is extremely small. In this case, the
relative impact of the online warmstart can be less pronounced (it is
certainly strictly better still) over an arbitrary initialization of
L-BFGS. However, as we saw on our splice site recognition data,
on this extreme, the online
learning methods will typically struggle since they are usually quite
effective in fitting the data to moderate but not high accuracies (as
evident from their $1/\epsilon$ or $1/\epsilon^2$ convergence
rates). Overall, we find that even on these two extremes, the hybrid
approach is competitive with the better of its two components.

\subsection{Communication Cost Comparison with Previous Approaches}
\label{sec:commcomplexity}

In the previous section we discussed the computational complexity of
several techniques with an identical communication pattern: communication
of the entire weight vector once per pass. In
this section we contrast our approach with techniques that use
other communication patterns. We focus mainly on communication
cost since the computational cost is typically the same as for
our algorithm, or the communication dominates the computation.

Since modern network switches are quite good at isolating
communicating nodes, the most relevant communication cost is the
maximum (over nodes) of the communication cost of a single node.

Several variables (some of them recalled from the previous section)
are important:
\begin{enumerate}
\item $m$ the number of nodes.
\item $n$ the total number of examples across all nodes.
\item $s$ the number of nonzero features per example.
\item $d$ the parameter dimension.
\item $T$ the number of passes over the examples.
\end{enumerate}
In the large-scale applications that are subject of this paper,
we typically have $s\ll d \ll n$, where both $d$ and $n$ are
large (see Section~\ref{sec:datasets}).

The way that data is dispersed across a cluster is relevant in much of
this discussion since an algorithm not using the starting format must
pay the communication cost of redistributing the data. We assume
the data is distributed across the nodes uniformly according to an
example partition, as is common.

\begin{table}
\caption{Communication cost of various  learning algorithms.
  Here $n$ is the number of examples, $s$ is the number of nonzero
  features per example, $d$ is the number of
  dimensions, $T$ is the number of times the algorithm examines
  each example, and $b$ is the minibatch size (in minibatch algorithms).}%
\centering\footnotesize%
\begin{tabular}{l|c}
Algorithm & Per-node communication cost\tabularnewline
\hline
Bundle method \citep{TeoLeSmVi07}& $O(dT_\text{bundle})$\tabularnewline
Online with averaging \citep{McDonaldHaMa2010,HallGiMa2010} & $O(dT_\text{online})$\tabularnewline
Parallel online \citep{HsuKaLaSm11} & $O(ns/m + n T_\text{online'})$\tabularnewline
Overcomplete online with averaging \citep{ZinkevichWeSmLi2010} & $O\left(ns+d\right)$\tabularnewline
Distr.~minibatch (dense) \citep{DekelGiShXi10,AgarwalDu2011} & $O\left(dT_\text{mini}n/b\right)=O\left(dT_\text{mini}\sqrt{n}\right)$\tabularnewline
Distr.~minibatch (sparse) \citep{DekelGiShXi10,AgarwalDu2011} & $O\left(bsT_\text{mini}n/b\right) = O\left( ns T_{\text{mini}} \right)$\tabularnewline
Hybrid online+batch & $O(dT_\text{hybrid})$
\end{tabular}
\end{table}

The per-node communication cost of the hybrid algorithm is $\Theta(d
T_\text{hybrid})$ where $T_\text{hybrid}$ is typically about $15$ to maximize test accuracy in
our experiments. Note that the minimum possible communication cost is
$\Theta(d)$ if we save the model on a single machine. There is no
communication involved in getting data to workers based on the data
format assumed above.
%
%Sample complexity arguments suggest (there are
%many caveats) $d \simeq \Theta(n)$ examples are needed for a
%meaningful statistical error, implying that the max node communication
%is $\Theta(n T_\text{hybrid})$.
%
An important point here is that every node has a
communication cost functionally smaller than the size of the dataset,
because there is no dependence on $ns$.

Similar to our approach, \citet{TeoLeSmVi07} propose a parallel batch
optimization algorithm (specifically, a bundle method) using the MPI
implementation of AllReduce. This approach arrives at an accurate
solution with $O(d T_{\text{bundle}})$ communication per node. Our
approach improves over this in several respects. First, as
Figure~\ref{fig:testerror} demonstrates, we obtain a substantial boost
thanks to our warmstarting strategy, hence in practice we expect
$T_{\text{bundle}} > T_\text{hybrid}$. The second distinction is in
the AllReduce implementation. Our implementation is well aligned with
Hadoop and takes advantage of speculative execution to mitigate the
slow node problem. On the other hand, MPI assumes full control over
the cluster, which needs to be carefully aligned with Hadoop's
MapReduce scheduling decisions, and by itself, MPI does not provide
robustness to slow nodes.

Batch learning can also be implemented using
MapReduce on a Hadoop cluster~\citep{ChuKiLiYuBrNgOl07},
for example in
the Mahout project.\footnote{\texttt{http://mahout.apache.org/}}
Elsewhere it has been noted that MapReduce is not well suited for
iterative machine learning
algorithms~\citep{LowGoKyBiGuHe10,MatMoTaAnJuMuMiScIo11}.  Evidence of
this is provided by the Mahout project itself, as their implementation
of logistic regression is not parallelized.  Indeed, we observe
substantial speed-ups from a straightforward substitution of MapReduce
by AllReduce on Hadoop.  It is also notably easier to program with
AllReduce, as code does not require refactoring.

The remaining approaches are based on online convex optimization.
\citet{McDonaldHaMa2010} and \citet{HallGiMa2010} study the approach
when each node runs an online learning algorithm on its examples and
the results from the individual nodes are averaged. This simple method
is empirically rather effective at creating a decent solution.  The
communication cost is structurally similar to our algorithm $\Theta(d
T_{\text{online}})$ when $T_{\text{online}}$ passes are done.
However, as we saw empirically in Figure~\ref{fig:testerror} and
also briefly argued theoretically in Section~\ref{sec:compcomplexity},
$T_{\text{online}} > T_\text{hybrid}$.
%Also, we have observed that
%non-uniform averaging approaches can provide a significant performance
%boost (see Table~\ref{tab:average}).

Similarly to these, \citet{ZinkevichWeSmLi2010} create an overcomplete
partition of the data and carry out separate online optimization on each
node followed by global averaging. Our experiments show that this
algorithm can have competitive convergence (e.g., on display
advertising data), but on more difficult optimization problems it can
be much slower than the hybrid algorithm we use here (e.g., on splice
site recognition data).  This approach also involves deep replication
of the data---for example, it may require having 1/4 of all the
examples on each of 100 nodes.  This is generally undesirable with
large datasets. The per-node communication cost is $\Theta(n s
T_{\text{rep}} / m + d)$ where $T_{\text{rep}}$ is the level of
replication and $m$ is the number of nodes.  Here, the first term
comes from the data transfer required for creating the overcomplete
partition whereas the second term comes from parameter averaging.
Since $T_{\text{rep}}/m$ is often a constant near $1$ ($0.25$ was
observed by~\citealp{ZinkevichWeSmLi2010}, and the theory predicts
only a constant factor improvement), this implies the communication
cost is $\Theta(n s)$, the size of the dataset.

Other authors have looked into online mini-batch optimization
\citep{DekelGiShXi10,AgarwalDu2011}. The key problem here is the
communication cost.  The per-node communication cost is
$\Theta(T_{\text{mini}} d n / b)$ where $b$ is the minibatch size
(number of examples per minibatch summed across all nodes),
$T_{\text{mini}}$ is the number of passes over the data, $n/b$ is the
number of minibatch updates per pass and $d$ is the number of
parameters. Theory suggests $b\le\sqrt{n}$, implying
communication costs of $\Theta(T_\text{mini} d \sqrt{n})$.  While for
small minibatch sizes $T_\text{mini}$ can be quite small (plausibly
even smaller than $1$), when $d$ is sufficiently large, this
communication cost is prohibitively large.  This is the reason for the
slow performance of mini-batched optimization that we observed in our
experiments.  Reworking these algorithms with sparse parameter
updates, the communication cost per update becomes $bs$ yielding an
overall communication cost of $\Theta(T_{\text{mini}} n s)$, which is
still several multiples of the dataset size.  Empirically, it has also
been noted that after optimizing learning rate parameters, the optimal
minibatch size is often $1$~\citep{HsuKaLaSm11}.

Another category of algorithms is those which use online learning with
a \emph{feature based} partition of examples~\citep{HsuKaLaSm11, DeaCorMon12}.
The advantage of this class of algorithms is that they can scale to a very large 
number of parameters, more than can be fit in the memory of a single machine.
Several families of algorithms have been tested in~\citep{HsuKaLaSm11}
including delayed updates, minibatch, second-order minibatch,
independent learning, and backpropagation.  The per-node communication
costs differ substantially here.  Typical communication costs are
$\Theta(ns/m + n T_\text{online'})$ where the first term is due to
shuffling from an example-based format, and the second term is for the
run of the actual algorithm. The complexity of our approach is
superior to this strategy since $n \gg d$.

\section{Discussion}
\label{sec:discuss}

We have shown that a new architecture for parallel learning based on a
Hadoop-compatible implementation of AllReduce can yield a combination
of accurate prediction and short training time in an easy
programming style. The hybrid algorithm we employ allows us to benefit
from the rapid initial optimization of online algorithms and the high
precision of batch algorithms where the last percent of performance
really matters. Our experiments reveal that each component of our
system is critical in driving the performance benefits we
obtain. Specifically, Table~\ref{tbl:mapreduce} and
Figure~\ref{fig:warmstart} show the performance gains resulting from
our use of AllReduce and the warmstart of the L-BFGS algorithm.
The effectiveness of our overall system, as compared to
the previous approaches, is confirmed in Figure~\ref{fig:marty}. Two
issues we do not discuss in this paper are the overheads of data
loading and node scheduling within Hadoop. These issues can indeed
affect the performance, but we found that they typically get amortized
since they are one-time overheads in the AllReduce approach as opposed
to per-iteration overheads in MapReduce. Nonetheless, improvements in
the scheduling algorithms can further improve
the overall performance of our system.

Our paper carefully considers various design choices affecting
the communication and computation speeds of a large-scale linear
learning system, drawing from and building upon the available
techniques in the literature. The resulting system enables the
training of linear predictors on datasets of size unmatched in
previous published works.

\bibliography{bib}

\end{document}